\documentclass[10pt,twocolumn,letterpaper]{article}

\usepackage{iccv}
\usepackage{times}
\usepackage{epsfig}
\usepackage{graphicx}
\usepackage{amsmath}
\usepackage{amssymb}
 
\usepackage{subfigure}
\usepackage{gensymb}
\usepackage{algorithm}
\usepackage{algorithmic}
\usepackage{bm}
\usepackage{multirow}

\usepackage[breaklinks=true]{hyperref}

\iccvfinalcopy 

\def\iccvPaperID{2456} 

\ificcvfinal\fi
\begin{document}

\title{Semantic-embedded
Unsupervised Spectral Reconstruction \\from Single RGB Images in the Wild}

\author{Zhiyu Zhu$^{1}$, Hui Liu$^{1}$, Junhui Hou$^{1}$\thanks{Corresponding author. This work was supported by the Hong Kong Research Grants Council under grant CityU 11219019.}, Huanqiang Zeng$^{2}$, Qingfu Zhang$^{1}$\\
$^{1}$ City University of Hong Kong, $^{2}$ Huaqiao University\\
{\tt\small zhiyuzhu2-c@my.cityu.edu.hk, hliu99-c@my.cityu.edu.hk, jh.hou@cityu.edu.hk,}\\{ \tt\small zeng0043@hqu.edu.cn, qingfu.zhang@cityu.edu.hk.}}
\maketitle
\ificcvfinal\thispagestyle{empty}\fi
\begin{abstract}
    This paper investigates the problem of reconstructing  hyperspectral (HS) images from single RGB images captured by commercial cameras, \textbf{without} using paired HS and RGB images during training. To tackle this challenge,  we propose a new lightweight and end-to-end learning-based framework. Specifically, on the basis of the intrinsic imaging degradation model of RGB images from HS images, we progressively spread the differences between input RGB images and re-projected RGB images from recovered HS images via effective unsupervised camera spectral response function estimation. To enable the learning without paired ground-truth HS images as supervision, we adopt the adversarial learning manner and boost it with a simple yet effective $\mathcal{L}_1$ gradient clipping scheme. Besides, we embed the semantic information of input RGB images to locally regularize the unsupervised learning, which is expected to promote pixels with identical semantics to have consistent spectral signatures. In addition to conducting quantitative experiments over two widely-used datasets for HS image reconstruction from synthetic RGB images, we also evaluate our method by applying recovered HS images from real RGB images to HS-based visual tracking. Extensive results show that our method significantly outperforms state-of-the-art unsupervised methods and even exceeds the latest supervised method under some settings. The source code is public available at \url{https://github.com/zbzhzhy/Unsupervised-Spectral-Reconstruction}.
\end{abstract}

\section{Introduction}
As hyperspectral (HS) images delineate more reliable and accurate 
spectrum information of scenes than traditional RGB images, 
they facilitate many vision-based applications, such as 
visual tracking \cite{xiong2020material,van2010tracking}, detection \cite{makantasis2015deep,pham2019airborne}, and segmentation \cite{ravi2017manifold,aytaylan2016semantic}.
However, it is costly to acquire HS images \cite{zhu2020hyperspectral}, which 
severely limits its wide deployment.

Without relying on expensive and specially designed hardware, recovering HS images from single RGB images via computational methods promises a convenient and low-cost manner, 
and  
recent deep neural network (DNN)-based methods have demonstrated the impressive abilities in 
solving such a highly ill-posed recovery problem. However, the majority of them have to be trained with 
paired RGB and HS images  \cite{li2020adaptive,zhang2020pixel,xiong2017hscnn}. Unfortunately,  it is non-trivial to collect a large number of such paired data via specially designed devices, e.g., well calibrated dual cameras. 
Although one can inversely synthesize RGB images from available HS images to form paired data for training, 
the huge gaps between the synthetic and real images may degrade the performance of a model trained with synthetic data when applied to real data. 



\begin{figure*}[t]
    \centering
    \includegraphics[width=0.9\textwidth]{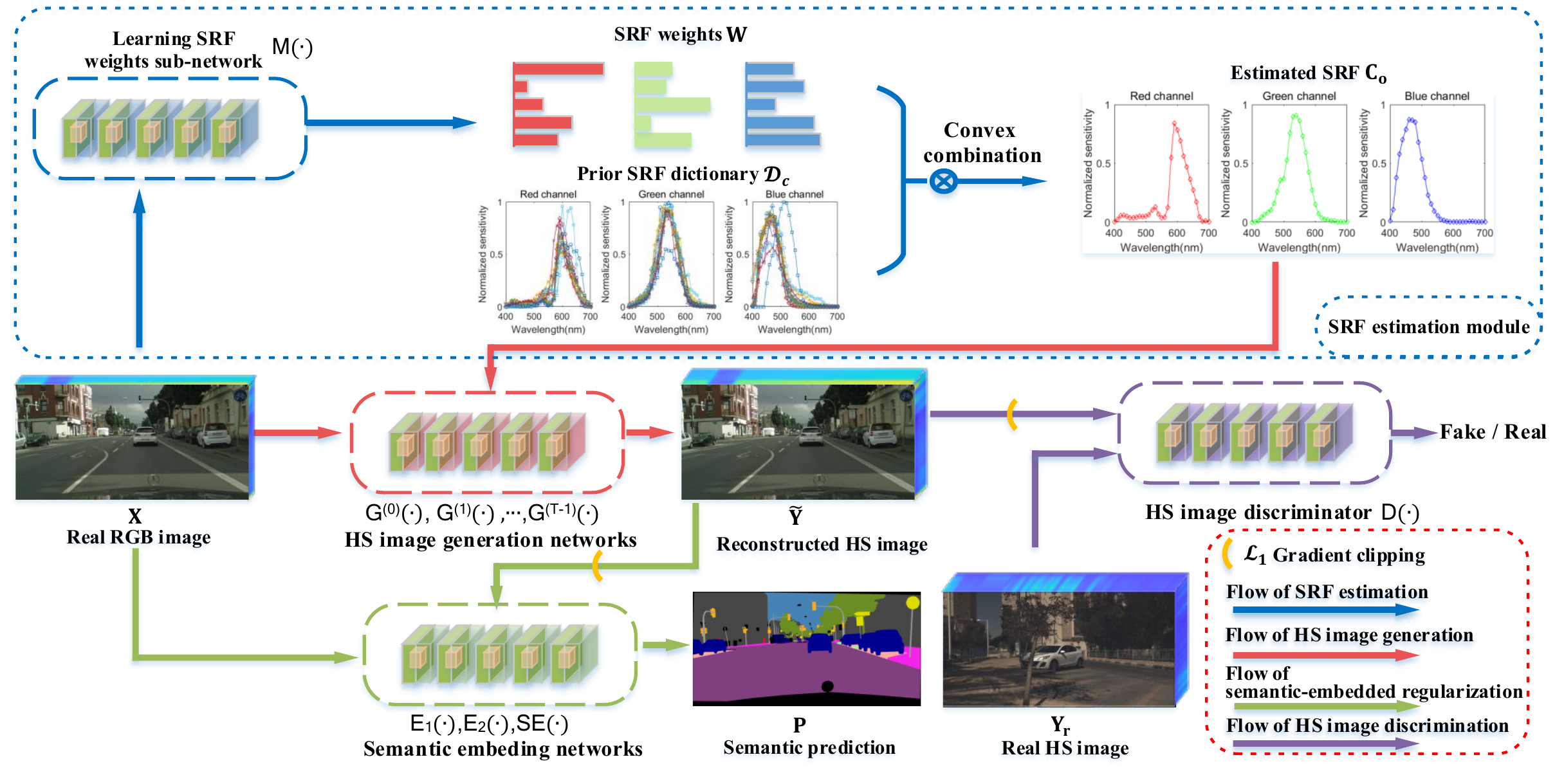}
	\vspace{-0.2cm}
    \caption{The flowchart of our \textbf{unsupervised} and \textbf{lightweight} learning-based framework. 
    \textbf{Unpaired} semantically-labeled RGB images in the wild and real HS images are used  for training. 
    \textbf{Note} that the two modules, i.e., the semantic-embedded regularization  and the $\mathcal{L}_1$ gradient clipping boosted discriminator, 
    are only used during training and not needed during testing. We refer readers to the \textit{Supplementary Material} for the detailed network architecture of each module.}
    \label{fig:flowchart}
	\vspace{-0.5cm}
\end{figure*}

To tackle the above issues, as illustrated in Fig. \ref{fig:flowchart}, we propose a lighweight, unsupervised, and end-to-end learning-based framework, which is capable of recovering HS images from single RGB images captured by commercial cameras \textit{without} using paired RGB and HS images during training. 
Specifically, based on the low-dimensional property of camera spectral response functions (SRFs), we first propose a prior-driven method to estimate the underlying camera SRFs of input RGB images by extracting their deep features.  
Then, we propose an imaging degradation model-aware HS image generation approach, in which an HS image is generated in a coarse-to-fine manner by progressively spreading the information of the difference between input RGB images and re-projected RGB images obtained by integrating recovered HS images via the estimated camera SRFs. Such a generation manner well adapts to this unsupervised scenario.  
To enable the meaningful learning of our method under the lack of paired ground-truth HS images, we make types of efforts:  
(1) \textbf{globally}, we employ the popular and powerful adversarial learning to enforce the distribution of generated HS images to be close to that of real HS images, Moreover, we propose $\mathcal{L}_1$ gradient clipping to stabilize and boost this learning manner;   
(2) \textbf{locally}, we embed the semantic information of scenes to encourage the pixels of reconstructed HS images with identical (resp. different) semantics to be similar (resp. dissimilar).   

The main contributions of this paper are summarized as follows: \vspace{-0.25cm}
\begin{enumerate}
    \itemsep-0.3em
    \item we propose an \textit{unsupervised} 
    HS image reconstruction framework from single RGB images in the wild; 
    \item  we propose imaging degradation model-aware 
    HS image generation and prior-driven 
    unsupervised estimation of camera spectral response functions;  
    \item we propose to  embed  scenes' semantic information efficiently to regularize the unsupervised learning;
    \item we propose a simple yet effective $\mathcal{L}_1$ gradient clipping strategy to stabilize and boost adversarial learning; and 
    \item   we introduce visual tracking-based quality evaluation of HS images recovered from real RGB images. 
\end{enumerate}

\section{Related Work}

\textbf{Spectral Recovery from Single RGB Images}. Based on the assumption that HS images lie in a low-dimensional subspace, many traditional methods explore the map between RGB images and subspace coordinates. For example, Nguyen \textit{et al.} \cite{nguyen2014training} leveraged an RGB white-balancing algorithm to normalize the scene illumination for the scene reflectance recovery.  Heikkinen \textit{et al.} \cite{heikkinen2018spectral} utilized a scalar-valued Gaussian process regression with an-isotropic or combination kernels to estimate the spectral subspace coordinates. Arad \textit{et al.} \cite{arad2016sparse} proposed a sparse coding-based method, which learns an over-complete dictionary of HS images to describe the novel RGB images. Then Aeschbacher \textit{et al.} \cite{aeschbacher2017defense} further improved it through introducing a shallow A+-based method \cite{timofte2014a+}. 

Owing to the impressive representation ability of DNNs, DNN-based methods have been proposed to solve the challenging problem of HS image reconstruction from single RGB images. For example, Xiong \textit{et al.} \cite{xiong2017hscnn} proposed HSCNN-D for the reconstruction of HS images from RGB images and compressive HS imaging. Shi \textit{et al.} \cite{shi2018hscnn+} further improved HSCNN-D by introducing residual blocks and dense connections with a cross-scale fusion scheme to facilitate the feature extraction process. 
Fu \textit{et al.} \cite{fu2018spectral} explored non-negative structured information and utilized multiple spare dictionaries to learn a compact basis representation for spectral reconstruction. To solve the uncertainty of camera SRFs, Berk \textit{et al.} \cite{kaya2019towards} trained a model to select different models, which are trained separately based on different SRFs. 
Li \textit{et al.} \cite{li2020adaptive} utilized both channel attention and spatial non-local attention for spectral reconstruction. Based on the assumption that pixels in an HS image belonging to different categories or spatial positions often require distinct mapping functions, Zhang \textit{et al.} \cite{zhang2020pixel}  proposed a pixel-aware deep function-mixture network, which learns reconstruction functions with different receptive fields and linearly mixes them up according to pixel-level weights. Aitor \textit{et al.} \cite{alvarez2017adversarial} applied a generative adversarial network to capture spatial semantics and map RGB to HS images. Yan \textit{et al.} \cite{yan2020reconstruction} introduced prior category information to generate distinct spectral data of objects via a U-Net-based architecture. 
Fu \textit{et al.} \cite{fu2018joint} developed an SRF selection block to retrieve the optimal response function for HS image reconstruction.  
Galliani \textit{et al.} \cite{galliani2017learned} utilized a densely connected U-Net-based architecture for HS images reconstruction. Most of the aforementioned methods have to be trained with ground-truth HS images of input RGB images. 
However, such a heavy demand for paired HS and RGB images is hard to satisfy in practice. 
Although only a few works \cite{zhang2020pixel,yan2020reconstruction} have been  aware of the necessity of object context for HS image recovery, they could not explicitly utilize semantic information.


\textbf{Generative Adversarial Networks (GANs)}. 
GANs have been widely adopted for image synthesis \cite{odena2017conditional,regmi2018cross} and reconstruction \cite{wang2018esrgan,kupyn2018deblurgan}With the emergency of GANs, the works on  improving GAN's stability and reducing the mode collapse have never stopped \cite{arjovsky2017wasserstein,heusel2017gans,miyato2018spectral,karras2017progressive,thanh2019improving}. Among of them, some works solve the problems through replacing loss functions, e.g.,  Wasserstein GANs (W-GANs) \cite{arjovsky2017wasserstein}, least squares GANs \cite{mao2017least}.
Others attempt to stabilize GANs' training via regularizing discriminator. For example, Gulrajani \textit{et al.} \cite{NIPS2017_892c3b1c} proposed weight-clipping and gradient penalty to stabilize the training of W-GANs. Miyato \textit{et al.} \cite{miyato2018spectral} proposed spectral normalization. 
However, these methods inevitably lose the discriminant ability, 
while enjoying the performance gains.


\section{Proposed Method}
\subsection{Problem Statement and Overview}
The physical degradation model of RGB images from HS images 
could be generally formulated as\vspace{-0.15cm}
 \begin{align}
    \label{Eq:degradation}
    \mathbf{X} = \mathbf{C}\mathbf{Y} + \mathbf{N}_y, \vspace{-0.2cm}
 \end{align}
 where $\mathbf{X} \in \mathbb{R}^{3\times hw}$ denotes the vectorial representation of an RGB image of spatial dimensions $h\times w$,  $\mathbf{Y} \in \mathbb{R}^{s\times hw}$ is the corresponding HS image with $s$  ($s\gg3$) spectral bands to be reconstructed, $\mathbf{C} \in \mathbb{R}^{3\times s}$ is the camera spectral response function (SRF), and $\mathbf{N}_y \in \mathbb{R}^{3\times hw}$ is the noise.
 
Apparently, one can design and train a DNN with paired RGB and HS images to 
learn the non-linear mapping from $\mathbf{X}$ to $\mathbf{Y}$. 
However,  
it is costly, laborious and inconvenient to collect massive such image pairs  
for training in practice.  Besides, due to different spatial resolution, field of views, and focal lengths, it is non-trivial to well register RGB and HS images captured by dual cameras. Alternatively, one may consider synthesizing RGB images from real HS images to form paired training data. However, 
the inevitable gap between real and synthetic RGB images may result in the DNN trained with synthetic data cannot be well generalized to real data. 

To this end, as shown in Fig. \ref{fig:flowchart}, we propose an \textit{unsupervised}, \textit{lightweight}, and \textit{end-to-end} learning-based HS image reconstruction framework from single RGB images captured by commercial cameras, where the corresponding ground-truth HS images of input RGB images are not required during training. Our framework is mainly composed of four modules: 

(1) \textbf{Degradation model-aware HS image generation}: 
Let $\widehat{\mathbf{Y}}$ be the recovered HS image from $\mathbf{X}$, and $\widehat{\mathbf{X}}$ be the re-projected RGB image by applying the camera SRF of $\mathbf{X}$ to $\widehat{\mathbf{Y}}$. According to Eq. (\ref{Eq:degradation}), if $\widehat{\mathbf{Y}}$ approximates the ground-truth one well, the difference between $\mathbf{X}$ and  $\widehat{\mathbf{X}}$ should be very small. 
Based on this intrinsic degradation relationship, we propose an 
HS image generation network, where a coarse HS image is first generated and then progressively refined by spreading useful information embedded in the difference between $\mathbf{X}$ and  $\widehat{\mathbf{X}}$. 
The design of our generation network elegantly adapts to such an unsupervised scenario.
See Section \ref{Sec:Generator}.

(2) \textbf{Prior-driven camera SRF estimation}: To realize the re-projection of the generation network, the camera SRFs of input RGB images have to be known. However, for RGB images in the wild, their SRFs are camera model-dependent and usually unavailable.    
Based on the fact that the SRFs of commercial cameras actually lie in a low-dimensional convex space 
(e.g.,  the first two principle components contain over 97 \% of total variance of SRFs of 28 different camera models) \cite{jiang2013space},  we propose a prior-driven SRF estimation network, in which based on extracted deep features from input RGB images, SRFs are estimated as the convex combination of the atoms of a pre-defined SRF dictionary. See Section \ref{Sec:SRF}.

(3) \textbf{$\mathcal{L}_1$ gradient clipping boosted adversarial learning}: To enable the learning of the generation network under the lack of paired ground-truth HS images, we adopt the adversarial learning manner, i.e., a discriminator is used to promote the distribution of generated HS images to approach that of real HS images.
Besides, due to the inherent possibility of gradient explosion \cite{thanh2019improving}, it is non-trivial to train GANs \cite{thanh2019improving,zhang2019self}. Inspired by traditional gradient clipping \cite{bengio2017deep},  we propose a simple yet effective strategy, namely $\mathcal{L}_1$ gradient clipping, to stabilize the training process and likewise boost performance, in which the magnitude of the gradient matrix from the discriminator is adaptively clipped with reference to the gradient of $\mathcal{L}_1$ loss function.
See Section \ref{Sec:L1GC}

\begin{figure*}[t]
    \centering
	\vspace{-0.6cm}
    \includegraphics[width=\textwidth]{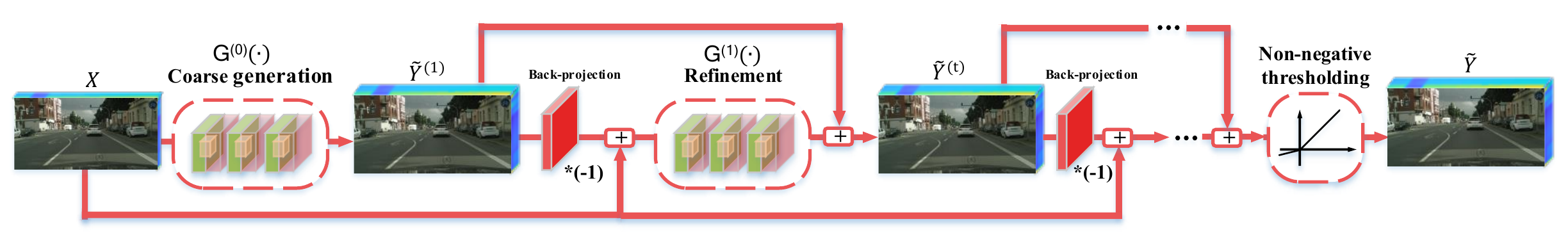}
	\vspace{-0.6cm}
    \caption{Illustration of the network architecture of the proposed degradation model-aware HS image generation. 
    }
    \label{fig:my_label}
	\vspace{-0.6cm}
\end{figure*}
(4) \textbf{Semantic-embedded regularization}: 
Owing to the rich spectrum information, the pixels of an HS image are discriminative 
\cite{camps2005kernel}, 
i.e., pixels corresponding to different (resp. identical) objects/materials always have distinct (resp. similar) spectral signatures \cite{7914752,chen2012hyperspectral}. By leveraging this unique property, 
we propose a semantic-embedded regularization network 
to further regularize the solution space of the unsupervised generation network for meaningful reconstruction,  i.e., during training, the generated HS images by the generation network are fed into a scene parsing/semantic segmentation network trained with the ground-truth semantic maps of the input RGB images to predict semantic maps. %
In contrast to the discriminator that \textit{globally} regularizes the distribution of HS images, this module is able to \textit{locally} regularize the pixels of generated HS images. See Section \ref{Sec:semantic}.  Note that RGB image-based semantic segmentation is a popular image analysis task and there are many publicly available datasets, and thus our method will not introduce additional costs. Moreover, the emerging weakly-supervised scene parsing methods \cite{wang2020self,chang2020weakly} will lessen the requirements of the ground-truth semantic annotations.


Note that our framework is end-to-end trained with the objective functions in Section \ref{Sec:objective}, and the two modules adversarial learning and semantic-embedded regularization are not needed during testing. In what follows, we will show the technical details of each module.

 \subsection{HS Image Generation}
 \label{Sec:Generator}
As shown in Fig. \ref{fig:my_label}, this module generates an HS image from an input RGB image in a coarse-to-fine fashion via the following three steps: 

\textbf{Coarse generation}. We first learn a coarse HS image denoted as $\mathbf{Y}^{(1)}$ using an efficient sub-CNN $\mathsf{G}^{(0)}(\cdot)$: 
\vspace{-0.2cm}
\begin{equation}
\begin{aligned}
    \widetilde{\mathbf{Y}}^{(1)} = \mathsf{G}^{(0)}(\mathbf{X}).
\end{aligned}
\end{equation}
 

\textbf{Progressive refinement}. 
We then progressively refine $\mathbf{Y}^{(1)}$ via a multi-stage structure, and at the $t$-th ($1\leq t\leq T-1$) stage, the refinement is carried out as 
\vspace{-0.2cm}
\begin{equation}
\begin{aligned}
    \widetilde{\mathbf{Y}}^{(t+1)} = \widetilde{\mathbf{Y}}^{(t)} + \mathsf{G}^{(t)}(\mathbf{C}_o \widetilde{\mathbf{Y}}^{(t)} - \mathbf{X}),
\end{aligned}\vspace{-0.2cm}
\end{equation}
where $\mathsf{G}^{(t)}(\cdot)$ stands for the sub-CNN at the $t$-th stage, and $\mathbf{C}_o\in\mathbb{R}^{3\times s}$ is the predicted camera SRF explained in Section \ref{Sec:SRF}.  
Note that $\mathsf{G}^{(1)}(\cdot)$, $\cdots$,   $\mathsf{G}^{(T-1)}(\cdot)$ have the same network architecture equipped with different parameters, and the network architecture of $\mathsf{G}^{(0)}(\cdot)$ is different from that of $\mathsf{G}^{(t)}(\cdot)$ because $\mathsf{G}^{(t)}(\cdot)$ perceives error maps, while $\mathsf{G}^{(0)}(\cdot)$ takes RGB images as input. 



\textbf{Non-negative thresholding}. Being aware that the pixel intensity of a physical meaningful HS image should be non-negative,  
we finally employ the  LeakyReLU, which does not obstruct gradient propagation, to guarantee the non-negativity of generated HS images: 
\begin{equation}
\begin{aligned}
    \widetilde{\mathbf{Y}} = \mathsf{LeakyReLU}\left(\widetilde{\mathbf{Y}}^{(T)}\right).
\end{aligned}\vspace{-0.3cm}
\end{equation}
where $\widetilde{\mathbf{Y}}\in\mathbb{R}^{s\times hw}$ is the generated HS image.


 \subsection{Camera SRF Estimation} 
 \label{Sec:SRF}

 Denote by $\mathcal{D}_c = \{ \mathbf{C}_1, \mathbf{C}_2, \cdots, \mathbf{C}_N \}$ a pre-defined  dictionary consisting of $N=28$ commonly-used camera SRFs \cite{jiang2013space}. We first estimate an intermediate SRF  $\widehat{\mathbf{C}} \in \mathbb{R}^{3\times s}$ as 
 \vspace{-0.2cm}
 \begin{align}
     &\widehat{\mathbf{C}}(i) = \sum_{j=1}^{N} \widehat{\mathbf{w}}_i(j) \times \mathbf{C}_j (i),~~(i=1,~2,~3),\nonumber \\
    & {\rm with}~~ \widehat{\mathbf{w}}_i=\mathcal{S}_{max}(\textbf{W}(i)), ~~ \mathbf{W}=\mathsf{M}(\mathbf{X}),
 \end{align}
 where 
 $\mathsf{M}(\cdot)$ is a sub-CNN extracting deep features from $\mathbf{X}$ and outputs the weight matrix $\mathbf{W}\in\mathbb{R}^{3\times N}$;  $\widehat{\mathbf{C}}(i)$, $\mathbf{C}_j(i)$, and $\mathbf{W}(i)$ are the $i$-th row of $\widehat{\mathbf{C}}$, $\mathbf{C}_j$, and $\mathbf{W}$, respectively; $\widehat{\mathbf{w}}_i(j)$ is the $j$-th entry of $\widehat{\mathbf{w}}_i\in\mathbb{R}^{1\times N}$;  
 $\mathcal{S}_{max}(\cdot)$ denotes the softmax operator. 

 Besides, real RGB images by commercial cameras are usually adjusted to be visually-pleasing via white balance. That is, the red, green, and blue three channels are separately re-scaled to have the same average value via the widely-used Gray World white balance algorithm \cite{lam2003image}. Considering its pixel position-independent property, 
 we mimic this operation by re-scaling the SRF:
 \begin{align}
     \mathbf{C}_o(i) = &\frac{\sum_{j=1}^{N} \widetilde{\mathbf{W}}(i,j)}{\sum_{j=1}^{N}\sum_{k=1}^{3} \widetilde{\mathbf{W}}(k,j) / 3} \times \widehat{\mathbf{C}}(i),~~(i=1,~2,~3),\nonumber \\
    & {\rm with} ~~ \widetilde{\mathbf{W}}=\mathsf{Sigmoid}(\mathbf{W}),
 \end{align}
 where $\mathbf{C}_o(i)$ is the $i$-th row of the finally estimated camera SRF $\mathbf{C}_o \in\mathbb{R}^{3\times s}$, $\mathsf{Sigmoid}(\cdot)$ is the sigmoid function, and $\widetilde{\mathbf{W}}(k,j)$ denotes the $(k, j)$-th entry of 
 matrix $\widetilde{\mathbf{W}}\in\mathbb{R}^{3\times N}$. 
 The use of the sigmoid function is able to avoid the cancellation when computing the the summation of  positive and negative values, and also maintains the original order. 
 \subsection{Adversarial Learning with Gradient Clipping}
 \label{Sec:L1GC}

 
It is commonly-known that training a good GAN is challenging, and the poor convergence may severely compromise 
 the final performance. Instead of using existing regularization methods \cite{miyato2018spectral,arjovsky2017wasserstein,NIPS2017_892c3b1c}, which may compromise the discriminant ability of the discriminator, we propose $\mathcal{L}_1$ gradient clipping  to stabilize the training process and likewise boost performance. 
 Specifically, the existing gradient clipping \cite{bengio2017deep}, which was originally designed for solving the gradient issue of training RNNs,  re-scales the intermediate gradient to a certain level, i.e., 
 , 
\begin{equation}
\begin{aligned}
    \mathcal{C}(\mathbf{G},\tau) =  \mathsf{Min}\left(1, \frac{\tau}{ \|\mathbf{G} \|_F }\right) \cdot \mathbf{G},
\end{aligned}
\label{equ:GP}
\end{equation}
 where $\mathcal{C}(\cdot,\cdot)$ is the gradient clipping function for training, $\tau$ is  the gradient threshold, $\mathbf{G} \in \mathbb{R}^{c\times hw}$ is the gradient matrix to be regularized, and $\|\cdot\|_F$ is the Frobenius norm of a matrix. Although gradient clipping can regularize the gradient magnitude without changing discriminator parameters and weakening its discriminant ability, the threshold $\tau$ has to be carefully set.  
 Based on the observation that the classic $\mathcal{L}_1$ loss function can produce a gradient with a fixed $\mathcal{F}$-norm, which only varying with the input size, we modify Eq. (\ref{equ:GP}) and propose $ \mathcal{L}_1$ gradient clipping, which re-scales the gradient matrix as $\mathcal{C}(\mathbf{G},\frac{1}{\sqrt{c \times hw}})$.
 We apply this $\mathcal{L}_1$ gradient clipping to the gradient matrix back-propagated from the discriminator to the HS image generation network. We  experimentally validated the \textbf{impressive} performance of such a simple strategy in Table \ref{tab:gradient}.
 
\subsection{Semantic-embedded HS image Regularization} 
 \label{Sec:semantic}

We utilize existing scene parsing datasets as the source of input RGB images during training, which contain semantically-labeled  RGB images in pixel-wise. 
As it is usually complex and time-consuming to train a scene parsing network, 
we propose the following simple manner: (1) first utilizing a pre-trained scene parsing network \cite{wang2020deep} as the backbone, denoted as $\mathsf{E}_1(\cdot)$, to extract features denoted as $\mathbf{\Phi}_X$ from $\mathbf{X}$ and a plain sub-CNN, denoted as $\mathsf{E}_2(\cdot)$, to extract features denoted as $\mathbf{\Phi}_{Y}$ from $\widetilde{\mathbf{Y}}$; and (2) then fusing $\mathbf{\Phi}_X$ and $\mathbf{\Phi}_Y$ via another sub-CNN, denoted as $\mathsf{SE}(\cdot,\cdot)$,  to predict the semantic map 
$\textbf{P} \in \mathbb{R}^{M \times hw}$ with 
$M$ being the total number of semantic classes.
It is expected that via back-propagation, the generation network will be promoted to  
produce $\widetilde{\mathbf{Y}}$ which could be distinguished by the semantic-embedded network trained with ground-truth semantic maps of $\mathbf{X}$. We also apply $\mathcal{L}_1$ gradient clipping to stabilize the fluctuated gradient from the semantic-embedded network. 
 
 \subsection{Objective Functions}
 \label{Sec:objective}
Analogy to most GAN-based frameworks, 
we alternately train the generator (including HS image generation,  semantic-embedded regularization, and camera SRF estimation) and the discriminator. The objective function of the generator is
\begin{equation}
\begin{aligned}
    \mathcal{L}_G =  \mathcal{B}(\mathsf{D}(\widetilde{\mathbf{Y}}/\overline{y}),~1) 
    + \lambda_b \mathcal{S}_{bp}(\mathbf{C}_o,\widetilde{\mathbf{Y}},\mathbf{X}) + \lambda_s \mathcal{S}_1(\widetilde{\mathbf{Y}})&\\
     + \lambda_p \mathcal{S}_{pos}(\widetilde{\mathbf{Y}})+ \lambda_{sem} \mathcal{S}_{sem}(\mathbf{P},\mathbf{L}),&
\end{aligned}
\end{equation}
where $\lambda_b$, $\lambda_s$, $\lambda_p$, and $\lambda_{sem}$ are non-negative parameters to balance different terms, which 
are empirically set to $1e^{2}$, $1e^{1}$, $1e^{-2}$, and $1e^{0}$, respectively; 
$\overline{y}$ is the mean value of $\widetilde{\mathbf{Y}}$; $\mathcal{B}(\cdot)$ is the binary cross-entropy loss; $\mathsf{D}(\cdot)$ stands for the discriminator; the $\mathcal{S}_{bp}(\cdot,\cdot,\cdot)$ is the back-projection constraint, $ \mathcal{S}_1(\cdot)$ is the first-order smoothness constraint, $\mathcal{S}_{pos}(\cdot)$ is the positive-definite constraint, and $\mathcal{S}_{sem}(\cdot,\cdot)$ is the loss for the semantics embedding regularization, where are respectively defined as 
\vspace{-0.25cm}
\begin{align}
    \mathcal{S}_{bp}(\mathbf{C}_o,\widetilde{\mathbf{Y}},\mathbf{X}) = \frac{1}{3 \times hw} \left\|\mathbf{C}_o\widetilde{\mathbf{Y}} - \mathbf{X}\right\|_F^2,
\end{align}
     \vspace{-0.8cm}
\begin{align}
    \mathcal{S}_1(\widetilde{\mathbf{Y}}) =\frac{1}{(s-1) \times hw} \sum_{i=1}^{s-1}
    \left\|\widetilde{\mathbf{Y}}(i)-\widetilde{\mathbf{Y}}(i+1)\right\|_1 ,
\end{align}
\vspace{-0.5cm}
\begin{align}
    \mathcal{S}_{pos}(\widetilde{\mathbf{Y}}) = \left\|\widetilde{\mathbf{Y}} - \left\lvert\widetilde{\mathbf{Y}}\right\rvert\right\|_F^2,
\end{align}
 \vspace{-0.6cm}
\begin{align}
    \mathcal{S}_{sem}(\mathbf{P},\mathbf{L}) = \frac{1}{hw} \sum_{n=1}^{hw}-\mathbf{L}(n) \log(\mathbf{P}(n)),
    \vspace{-0.3cm}
\end{align}
 where $\lvert\cdot\rvert$ computes the absolute value of the input in element-wise, $\|\cdot\|_1$ is the $\ell_1$-norm of a vector, $\mathbf{P}(n) \in \mathbb{R}^{M \times 1}$ is the semantic prediction of the $n$-th pixel, and
 $\mathbf{L}(n)\in\mathbb{R}^{M \times 1}$ is the ground-truth of $\mathbf{P}(n)$. 
 The objective function of the discriminator is
\begin{equation}
\begin{aligned}
    \mathcal{L}_D =  \mathcal{B}(\mathsf{D}(\widetilde{\mathbf{Y}}/\overline{y}),~0) +\mathcal{B}(\mathsf{D}(\mathbf{Y}_r/\overline{y}_r),~1),
\end{aligned}
\end{equation}
where $\mathbf{Y}_r\in\mathbb{R}^{s\times hw}$ is a real HS image, 
$\overline{y}_r$ is the mean value of $\mathbf{Y}_r$. 
We normalize the $\widetilde{\mathbf{Y}}$ and $\mathbf{Y}_r$ respectively with their mean values 
to focus the reconstruction process much on the spectral angular of HS images. 
Besides, the proposed $ \mathcal{L}_1$ gradient clipping operation in Section \ref{Sec:L1GC} will be used to regularize the backward gradients.
\section{Experiments}
\begin{figure*}
	\centering
	\includegraphics[
	width=0.8\textwidth, height= 0.5\linewidth]{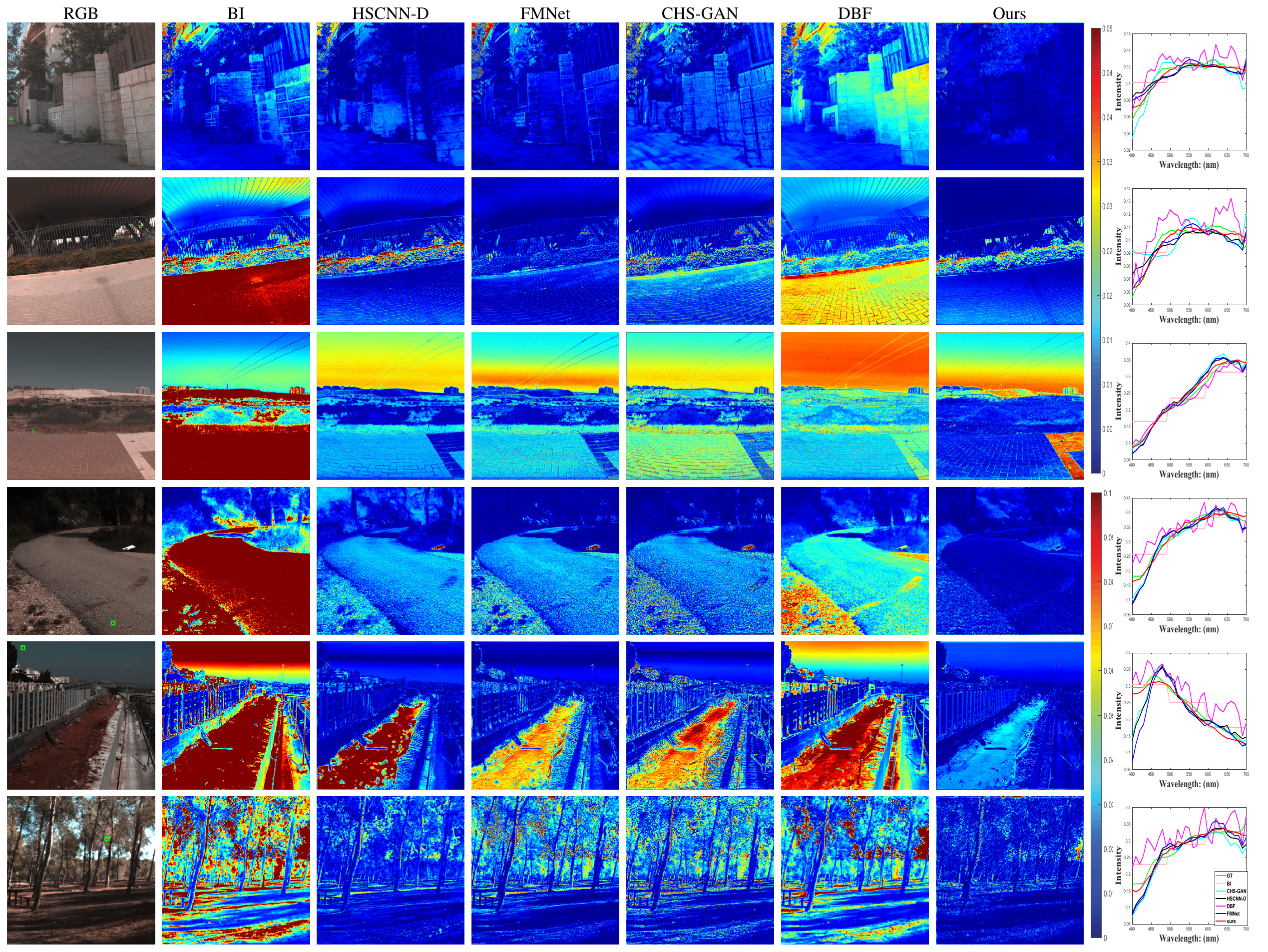}
	\caption{\label{fig:NTIRE_CAVE}Visualization of the error maps of different methods on 6 testing RGB images (the most left column) in the ICVL (the top three rows in 440, 590, and 640 nm) and NTIRE'20 (the bottom three rows in 490, 640, and 690 nm) datasets. Additionally, we also draw the spectral signatures of 6 pixels respectively selected from the 6 testing images 
	the selected pixels are marked by green rectangle in the RGB images.}
	\vspace{-0.5cm}
\end{figure*}

\textbf{Real HS image dataset}. The ICVL 
\cite{arad_and_ben_shahar_2016_ECCV} and NTIRE'20  
\cite{arad2020ntire} datasets  contain HS images of 31 spectral bands covering from 400 to 700 nm with interval of 10nm captured under outdoor daylight illumination. There are 201 HS images of spatial dimensions $1392 \times 1300$ and 460 HS images of spatial dimensions 482 $\times$ 512 in the ICVL and NTIRE'20 datasets, respectively.  

\textbf{Semantically-labeled real RGB image datasets}.  
The PASCAL-Context 
\cite{mottaghi_cvpr14} and Cityscapes 
\cite{Cordts2016Cityscapes} 
datasets  contain RGB images with pixel-wise 
categorical labels 
for semantic embedding.  PASCAL-Context contains 59 semantic classes, which is divided into 4,998/5,105 images for training and testing, respectively. The Cityscapes dataset is officially divided into 2,975/500/1525 images for training, validation, and testing, respectively. 

\textbf{RGB and HS video dataset for visual tracking}. 
The hyperspectral object tracking challenge (HOTAC) 
dataset \cite{xiong2020material} contains 50 RGB videos with each about 425 frames, where the bounding-boxes of target objects are  given. Besides, HOTAC provides 50 HS videos with target objects labeled with bounding-boxes, and the HS videos capture approximately the same scenes as the RGB videos\footnote{Although the RGB and HS videos have been registered, 
there is still a large disparity between them.}. 

\textbf{Compared methods}. We compared with 5 methods, including 1 baseline, i.e., the bicubic interpolation (BI) over the spectral dimension, and 4 most recent DNN-based methods, i.e., CHS-GAN \cite{alvarez2017adversarial}, DBF \cite{fubara2020rgb}, HSCNN-D \cite{shi2018hscnn+}, and FM-Net \cite{zhang2020pixel}. Note that CHS-GAN, HSCNN-D, and FM-Net are supervised and have to be trained with paired HS and RGB images, and DBF is unsupervised.
\begin{table*}
	\centering
	\caption{\label{tab:results1} Quantitative comparisons of different methods on the ICVL dataset. ``$\uparrow$ (resp. $\downarrow$)" indicates that the larger (resp. smaller), the better.  For \# Params and \# FLOPs, the smaller, the more compact and efficient. 
	}
\resizebox{0.8\textwidth}{!}{
	\begin{tabular}{l|cc|ccc|cccc}
		\hline
		\hline
		\textbf{Methods}  &Paired data&Real RGB & \# Params& \# FLOPs& Running time (ms) & PSNR $\uparrow$ & ASSIM $\uparrow$& SAM $\downarrow$ \\
		\hline
		BI		                           &$\times$&$\checkmark$ &--	&--&--	    &25.96 &0.911 &19.18 \\
		HSCNN-D	\cite{shi2018hscnn+} &$\checkmark$&$\times$ &3.61 M&5.22 T &5184.75ms &38.15 &0.992 &3.28   \\
	    FM-Net 	\cite{zhang2020pixel}  &$\checkmark$&$\times$ &11.79 M &17.07 T &4780.41ms &40.26 &0.993 &2.65 \\
		CHS-GAN \cite{alvarez2017adversarial}  &$\checkmark$&$\times$ &1.09M &0.91T &505.70ms &38.03 &0.988 &3.34  \\
		DBF \cite{fubara2020rgb} &$\times$&$\times$ &1.13M	&1.02T &531.25ms	 &30.75 &0.940 &14.64 \\
		Ours                                &$\times$ &$\checkmark$ &0.58M  &0.93T &561.90 ms &37.83 &0.991 &4.01 \\
		\hline
		\hline
	\end{tabular}}
\end{table*}

\begin{table*}
	\centering
	\vspace{-0.2cm}
	\caption{\label{tab:results2} Quantitative comparisons of different methods on the NTIRE'20 dataset. 
	}
\resizebox{0.8\textwidth}{!}{
	\begin{tabular}{l|cc|ccc|ccc }
		\hline
		\hline
		\textbf{Methods}  &Paired data&Real RGB & \# Params& \# FLOPs& Running time  & PSNR $\uparrow$ & ASSIM $\uparrow$& SAM $\downarrow$ \\
		\hline
		BI		                        &$\times$ &$\checkmark$ &--	&--&--	    &25.91 &0.876 &23.00   \\
		HSCNN-D	\cite{shi2018hscnn+}&$\checkmark$ &$\times$&3.61 M&0.890 T&1003.38 ms &31.45 &0.919 &10.76   \\
	    FM-Net 	\cite{zhang2020pixel} &$\checkmark$&$\times$ &11.79 M &2.955 T&1066.40 ms &34.25 &0.968 &9.11   \\
		CHS-GAN \cite{alvarez2017adversarial}&$\checkmark$&$\times$ &1.09 M	&0.15T&87.74 ms   & 30.63 &0.906 &15.38   \\
		DBF \cite{fubara2020rgb} &$\times$&$\times$ &1.13 M	&0.17T&85.91 ms   &31.51 &0.891 &16.81   \\
		Ours                       &$\times$&$\checkmark$ &0.58 M  &0.16T&134.81 ms &37.53 &0.979 &5.97  \\
		\hline
		\hline
	\end{tabular}}
\end{table*}

\subsection{Evaluation on Synthetic RGB Images}
\label{subsec:synthetic exp}

\textbf{Experiment settings}. 
Although there is a gap between synthetic and real data, we still conducted experiments on synthetic data as done in existing works \cite{zhang2020pixel}, \cite{shi2018hscnn+} to  have a quantitative and intuitive understanding of our framework.

We selected the last 20 HS images of ICVL to form the test dataset, and the remaining ones as the training dataset. All the HS images of NTIRE'20 were used as the test dataset.  
For fair comparisons, we applied the same dictionary of camera SRFs in Section \ref{Sec:SRF} to the training HS images to synthesize RGB images, generating paired training data, which were used to train
 HSCNN-D \cite{shi2018hscnn+}, FM-Net \cite{zhang2020pixel}, and CHS-GAN \cite{alvarez2017adversarial} with the paired synthetic RGB and HS images.
Following the original implementation \cite{fubara2020rgb}, we first trained DBF to learn a basis function 
and then trained it on the synthetic RGB images with the 
learnt basis function. The proposed method was trained with PASCAL-Context and the aforementioned training HS images from ICVL as RGB and HS image sources, respectively, with the batch size of 24. The number of stages $T$ was set to $4$. We adopted the ADAM \cite{kingma2014adam} optimizer for optimization of the generation network, the discriminator, and the camera SRF estimation network. The exponential decay rates were set as $\beta_1 = 0.9$ and $\beta_2 = 0.999$ for the first and second moment estimates, respectively. We also adopted the TTUR scheme \cite{heusel2017gans} to set the learning rate of the SRF estimation network, HS image generation network, and the HS image discriminator to $1e^{-4}$, $1e^{-4}$, and $5e^{-4}$, respectively. Meanwhile, for the  semantic embedding module, we fixed $\mathsf{E}_1(\cdot)$ and trained  $\mathsf{E}_2(\cdot)$ and $\mathsf{SE}(\cdot)$ using the SGD optimizer with the base learning rate, the momentum, and the weight decay equal to $1e^{-3}$, 0.9, and $1e^{-4}$, respectively. The poly learning rate policy with the power of 0.9 is used for dropping the learning rate \cite{wang2020deep}. The proposed method is trained 50 epochs with 208 iterations each epoch. We tested all the methods with the same protocols on the synthetic RGB images generated from the widely-used camera SRF--- Nikon D700. To quantitatively compare different methods, we adopted 3 commonly-used metrics, i.e., Peak Signal-to-Noise Ratio (PSNR), Average Structural Similarity Index (ASSIM) \cite{wang2002universal}, and Spectral Angle Mapper (SAM) \cite{yuhas1992discrimination}. We reported all the methods with the training results of the last epoch.

\textbf{Quantitative and visual results}. 
Table \ref{tab:results1} lists the results of different methods on ICVL,
where it can be seen that our method significantly outperforms the unsupervised DBF and achieves comparable performance to the supervised HSCNN-D and CHS-GAN, which are credited to the carefully designed framework. But the gap between our method and  FM-Net --- the most recent supervised method also indicates the potential of this topic.    
However, as listed in Table \ref{tab:results2}, our method even exceeds FM-Net on NTIRE'20,  
which somewhat demonstrates better generalization ability.
As visualized in Fig. \ref{fig:NTIRE_CAVE}, 
we can see that 
the error between the reconstructed spectral band and the corresponding ground-truth one is much smaller than those of compared methods. Such an advantage may be benefit from the semantic information of the PASCAL-Context dataset containing the class of sky.  Besides, the spectral signatures of selected pixels are closer to ground-truth ones. 
Finally, 
the proposed method consumes much less time than HSCNN-D and FM-Net.

\begin{figure}
	\centering
	\subfigure[distance plot]{\includegraphics[width = 0.49\linewidth]{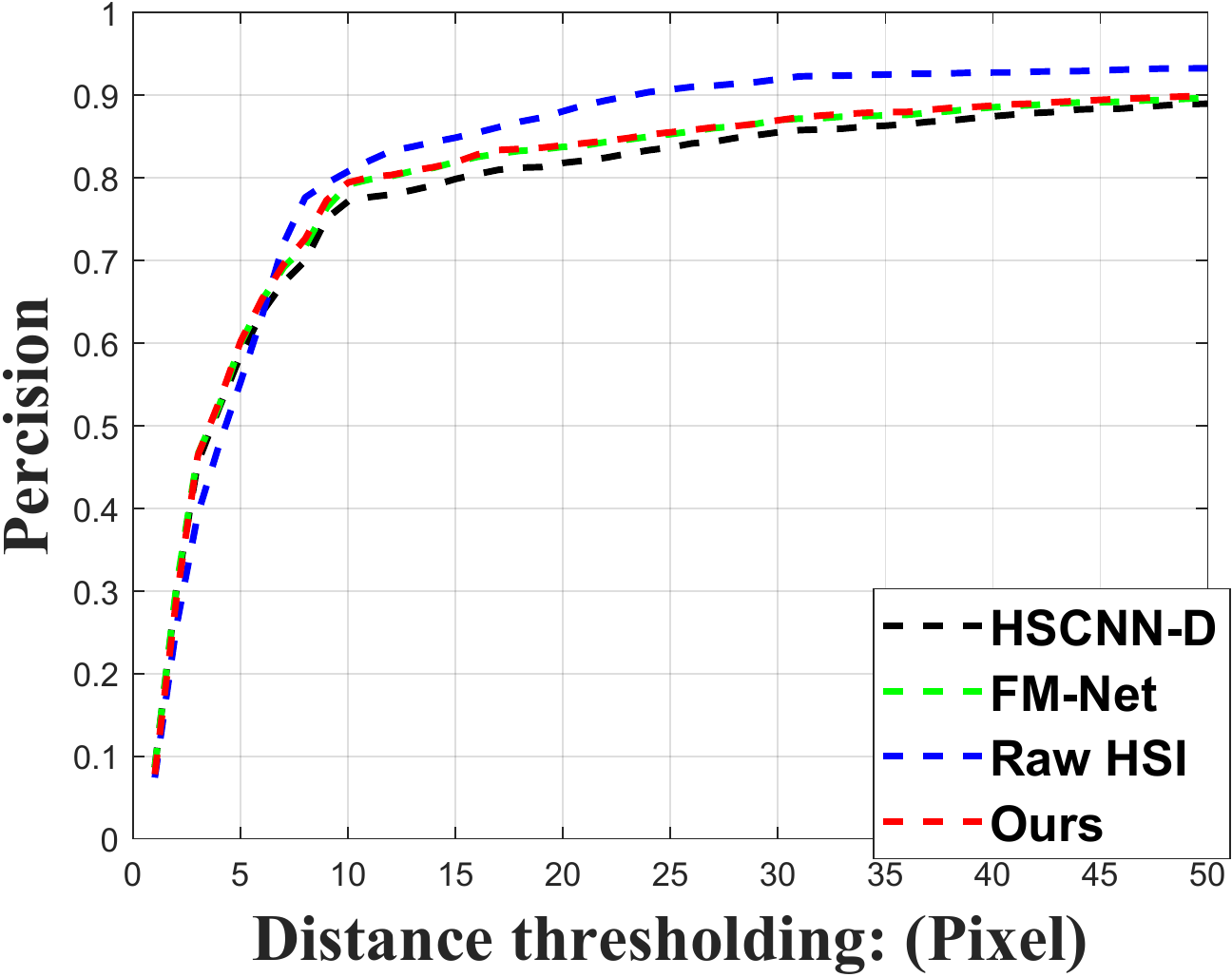}}
	\subfigure[success plot]{\includegraphics[width = 0.49\linewidth]{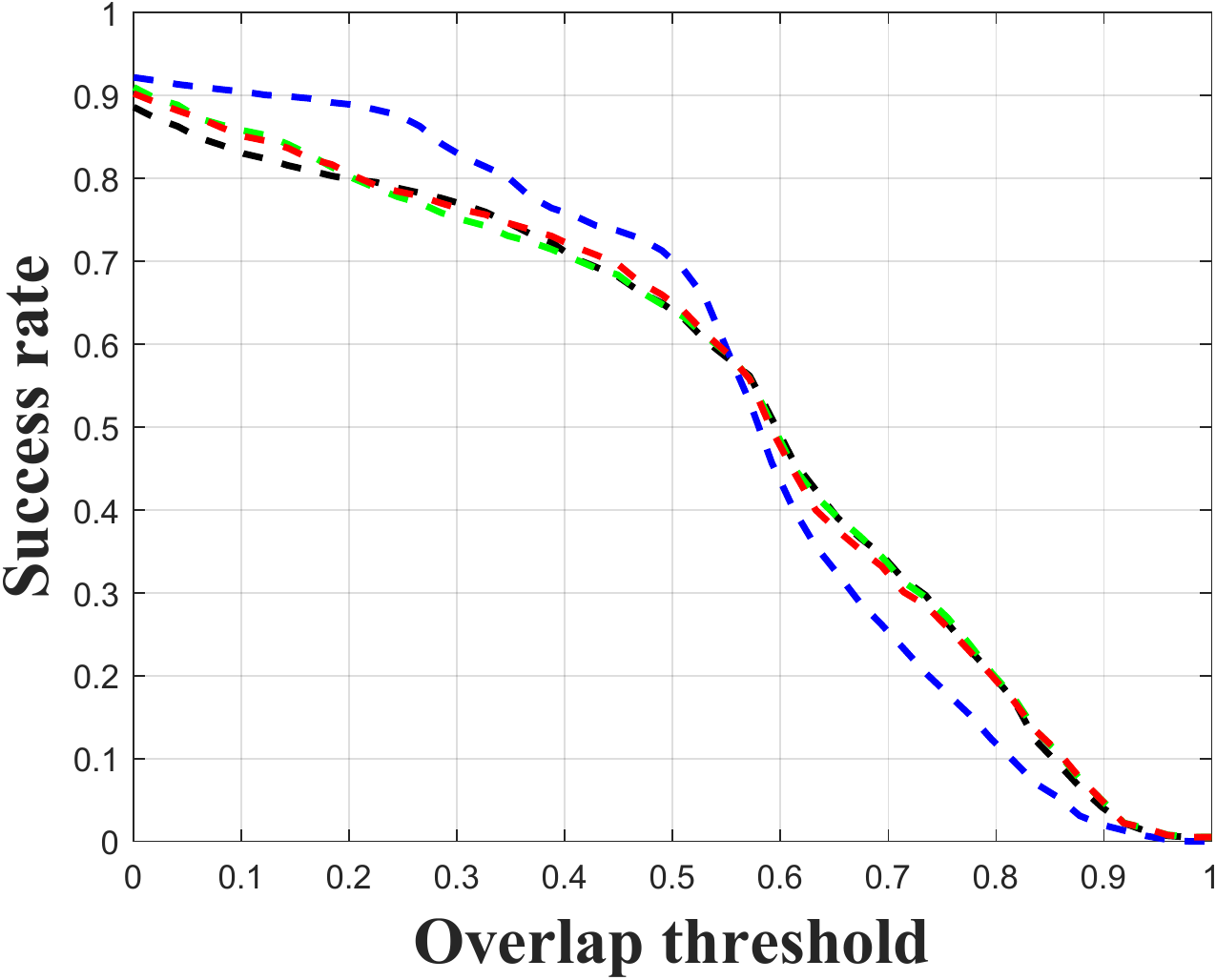}}
	\caption{\label{fig:tracking} Comparison of different methods evaluated with the visual tracking task.}
\end{figure}

\begin{table*}
\caption{\label{tab:ablation2} Results of ablation studies on the ICVL dataset.
$\times$ indicates that the corresponding component was removed when training testing the proposed method. The bottom row corresponds to the complete model.
}
\centering
\begin{tabular}{cccccc|ccc}
	\hline
	\hline
	Adversarial learning& $\mathcal{S}_{bp}$ & $\mathcal{S}_{pos}$& $\mathcal{S}_1$ & $\mathcal{L}_1$ Gradient clipping & $\mathcal{S}_{sem}$ &      PSNR $\uparrow$ & ASSIM $\uparrow$& SAM $\downarrow$ \\
	\hline
   $\checkmark$    &$\times$        & $\times$   & $\times$    & $\times$	& $\times$  &14.47 &0.121 &63.34\\ 
   $\checkmark$    &$\checkmark$    &$\times$    & $\times$    & $\times$	& $\times$  &20.48 &0.781 &35.70 \\ 
   $\checkmark$    &$\checkmark$    &$\checkmark$& $\times$    & $\times$    & $\times$  &21.72 &0.783 &30.64 \\  
   $\checkmark$    &$\checkmark$    &$\checkmark$& $\checkmark$& $\times$	 & $\times$ &26.11 &0.882 &14.44 \\ 
   $\checkmark$    &$\checkmark$    &$\checkmark$& $\checkmark$& $\checkmark$ & $\times$ &34.51 &0.983 &7.04 \\  
   $\checkmark$    &$\checkmark$    &$\checkmark$& $\checkmark$& $\checkmark$ & $\checkmark$ &37.83 &0.991 &4.01\\ 
	\hline
	\hline
\end{tabular}
\end{table*}
\subsection{Visual Tracking-based Evaluation on Real RGB Images}

For real RGB images, the well registered ground-truth HS images are usually unavailable, making it impossible to quantitatively evaluate different methods, as done in Section \ref{subsec:synthetic exp}. 
Besides, due to the high spectral resolution, it is also difficult to visually compare the results of different methods. 
Considering that the rich spectrum information makes pixels of HS images discriminative, which will be benefit 
high-level vision tasks, 
we introduced the HS video-based visual tracking experiment to evaluate different HS image reconstruction methods. 
It is expected a better method will generate HS images that are closer to unknown ground-truth ones, which then produce high tracking accuracy.


\textbf{Experiment settings}. 
We compared 
with 
HSCNN-D and FM-Net, which always achieve the top two performance among all the compared methods in the previous scenario. For all methods, we directly adopted the trained models in Section \ref{subsec:synthetic exp} \textbf{without} additional training to reconstruct the 10 randomly selected RGB videos from HOTAC  into HS videos, which were then fed to a pre-trained  HS video-based tracker named MHT \cite{xiong2020material} to realize tracking.
Besides, we also performed the MHT tracker on the 10 HS videos provided by HOTAC, corresponding to the 10 selected RGB videos. The tracking performance of such a setting, denoted as ``Raw", could be thought of as the upper-bound of the reconstruction methods.  
We quantitatively measured the tracking performance by using Average Precision (AP) among all the distance threshold, AUC, and location error, three commonly-used metrics in visual tracking. 

\textbf{Quantitative results}. Table \ref{tab:tracking} and Fig. \ref{fig:tracking} show the results, where it can be seen that our method even produces better tracking performance than the two supervised methods FM-Net and HSCNN-D, and as expected, the performance of all the three computation-based reconstruction methods is lower than that of ``Raw". 
Please refer to our \textit{Video Demo} for more visual results.

\begin{table}
	\centering
	\caption{\label{tab:tracking} Comparison of different methods on visual tracking.}
\resizebox{1.0\linewidth}{!}{
	\begin{tabular}{ c |ccc }
	\hline
	\hline
	Methods & AP (\%) $\uparrow$ & AUC $\uparrow$ & Location error $\downarrow$\\
	\hline
		 HSCNN-D \cite{shi2018hscnn+}  &78.65 &0.5235 &17.4358  \\
		 FM-Net \cite{zhang2020pixel} &80.03 &0.5268 &16.3825   \\
		 Ours    &80.23 &0.5272 &16.4884  \\
		 Raw    &83.21 &0.5360 &14.7835 \\
		\hline
		\hline
	\end{tabular}
	}
\end{table}
\begin{table}
	\centering
	\caption{\label{tab:gradient} Comparison of different regularization methods for boosting the training of GANs. 
	}
	\begin{tabular}{ c |ccc }
	\hline
	\hline
	Methods & PSNR $\uparrow$ & SSIM $\uparrow$ & SAM $\downarrow$\\
	\hline
		 1-GP \cite{NIPS2017_892c3b1c}  & 28.15 & 0.944 & 12.60  \\
		 0-GP \cite{thanh2019improving} & 33.33 & 0.984 & 6.29   \\
		 $\mathcal{L}_1$-GC (ours)   &37.83 &0.991&4.01  \\
		\hline
		\hline
	\end{tabular}
\end{table}
\begin{table}
	\centering
	\caption{\label{tab:dataset} Performance of our method trained with various RGB and HS image sources 
	The test data are 20 samples from ICVL.}
\resizebox{0.95\linewidth}{!}{
	\begin{tabular}{cc|ccc }
		\hline
		\hline
		RGB source  &HS source & PSNR $\uparrow$ & ASSIM $\uparrow$ & SAM $\downarrow$ \\
		\hline
		\hline
		Cityscapes     & NTIRE'20    &34.08 &0.9838 &6.61  \\
		Cityscapes     & ICVL          &37.68 &0.988  &5.03  \\
		PASCAL-Context & NTIRE'20    &35.87 &0.9837 &6.24  \\
		PASCAL-Context & ICVL          &37.83 &0.991  &4.01  \\
		\hline
		\hline
	\end{tabular}}
\end{table}
\subsection{Ablation Studies}

\textbf{Regularization terms}. We experimentally validated the effectiveness of the smooth constraint $\mathcal{S}_1(\cdot)$, the positive constraint $\mathcal{S}_{pos}(\cdot)$, the back-projection constraint $\mathcal{S}_{bp}(\cdot,~\cdot,~\cdot)$,  $\mathcal{L}_1$ gradient clipping, and the semantic-embedded module. As shown in Table \ref{tab:ablation2}, we can see that our framework benefits from each of these components. Among them, the back-projection regularization, semantic embedding module and $\mathcal{L}_1$ gradient clipping have the pivotal influence. Besides, 
only using traditional regularization terms such as $\mathcal{S}_{bp}(\cdot,~\cdot,~\cdot)$, $\mathcal{S}_{pos}(\cdot)$, and $\mathcal{S}_1(\cdot)$, the method produces very limited performance comparable to that of Bicubic interpolation, indicating that the unsupervised HS image reconstruction actually is a very challenging task.

\begin{figure}
    \centering
    \includegraphics[width=0.7\linewidth]{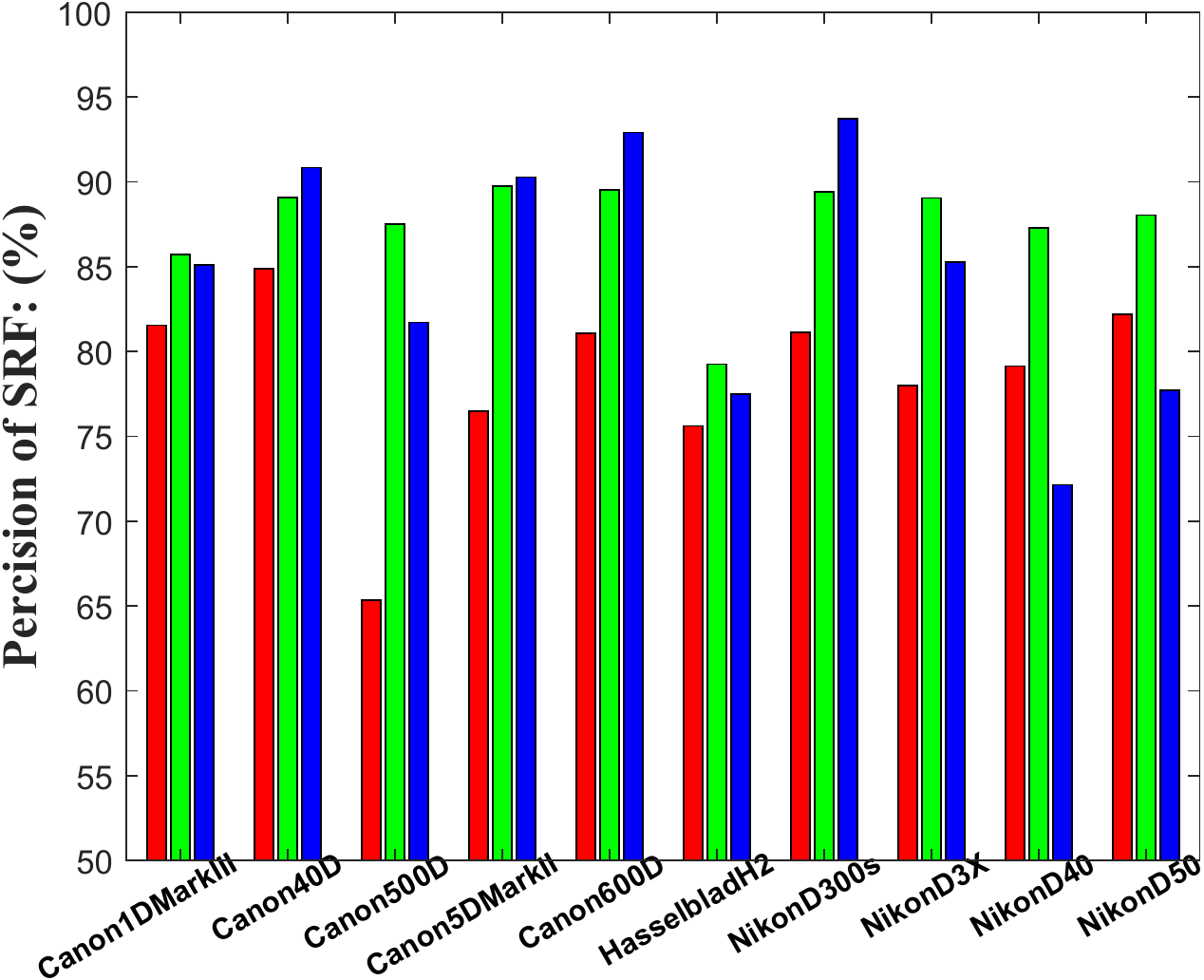}
    \caption{Reconstruction accuracy of our unsupervised prior-driven camera SRF estimation module. 
    }
    \label{fig:illustration}
\end{figure}

\textbf{Camera SRF estimation}. 
Even with supervision, camera SRF estimation of RGB images is a quite challenging task \cite{kaya2019towards}. To evaluate the effectiveness of our camera SRF estimation module, we randomly chose ten SRFs to generate synthetic RGB images from real HS images, which were then fed into our estimation module to estimate the SRFs. 
We calculated the precision of estimated SRFs using the AUC metric.  
As shown in Fig. \ref{fig:illustration},  the average precision of our estimation module achieves $77.79\%$. 
Note that our SRF estimation module is \textit{unsupervised}, i.e., the ground-truth camera SRFs were not employed as supervision during training. 

\textbf{$\mathcal{L}_1$ gradient clipping}. To show the advantage of the proposed $\mathcal{L}_1$ gradient clipping, we trained our framework with different gradient regularization methods \cite{thanh2019improving,NIPS2017_892c3b1c}, where we only changed the gradient regularization terms and remained the other modules unchanged. As listed in Table \ref{tab:gradient}, the proposed $\mathcal{L}_1$ gradient clipping exceeds the other gradient clipping methods to a large extent, demonstrating its superiority.

\textbf{Various training datasets}. To further explore the performance of our method on different datasets, we conducted experiments by training our method with different image datasets as training data. 
We tested all the trained models with the 20 samples of ICVL dataset. As listed in Table \ref{tab:dataset}, we can see that for the RGB image source, the performance of using PASCAL-Context is better than that of using Cityscapes, which may be credited to the more diverse samples contained in the PASCAL-Context dataset. 




\vspace{-0.1cm}
\section{Conclusion}
\vspace{-0.2cm}
We have presented a new unsupervised, lightweight, and end-to-end learning-based framework for the reconstruction of  HS images from single RGB images in the wild. Our framework does not need paired RGB and HS images for training. 
Owing the well-motivated reconstruction modules including prior-driven camera SRF estimation and imaging degradation model-aware generation, as well as the effective semantic embedding and $\mathcal{L}_1$ gradient clipping boosted adversarial learning   
for regularizing the learning of the framework without referencing ground-truth HS images, our framework achieves impressive performance, and especially our method even exceeds the most recent supervised reconstruction method under some settings. 


{\small
\bibliographystyle{ieee_fullname}
\bibliography{egbib}
}

\end{document}